\documentclass{article}

\PassOptionsToPackage{numbers, compress}{natbib}
\usepackage{xcolor}

\usepackage[final]{neurips_2020}

\usepackage[utf8]{inputenc} % allow utf-8 input
\usepackage[T1]{fontenc}    % use 8-bit T1 fonts
\usepackage{hyperref}       % hyperlinks
\usepackage{url}            % simple URL typesetting
\usepackage{booktabs}       % professional-quality tables
\usepackage{amsfonts}       % blackboard math symbols
\usepackage{nicefrac}       % compact symbols for 1/2, etc.
\usepackage{microtype}      % microtypography
\usepackage{amsmath}
\usepackage{amsfonts,amssymb}
\usepackage{algorithm}
\usepackage{algpseudocode}
\usepackage{titlesec}
\usepackage{graphicx}
\usepackage{multirow}
\usepackage{color, soul}

\title{Latent Template Induction with Gumbel-CRFs} 

\author{Yao Fu\textsuperscript{1,}\thanks{Work done during an internship at Alibaba DAMO Academy, in collaboration with PKU and Cornell.}~, Chuanqi Tan\textsuperscript{2}, Bin Bi\textsuperscript{2}, Mosha Chen\textsuperscript{2}, Yansong Feng\textsuperscript{3}, Alexander M. Rush\textsuperscript{4}  \\
\textsuperscript{1}ILCC, University of Edinburgh, \textsuperscript{2}Alibaba Group, \textsuperscript{3}WICT, Peking Univeristy, \textsuperscript{4}Cornell University\\
\texttt{yao.fu@ed.ac.uk}, 
\texttt{\{chuanqi.tcq; b.bi; chenmosha.cms\}@alibaba-inc.com} \\
\texttt{fengyansong@pku.edu.cn}, 
\texttt{arush@cornell.edu} \\
}

\begin{document}
% \begin{CJK*}{UTF8}{gbsn}

\maketitle

\begin{abstract}
Learning to control the structure of sentences is a challenging problem in text generation.
Existing work either relies on simple deterministic approaches or RL-based hard structures. 
We explore the use of structured variational autoencoders to infer latent templates for sentence generation using a soft, continuous relaxation in order to utilize reparameterization for training. 
Specifically, we propose a Gumbel-CRF, a continuous relaxation of the CRF sampling algorithm using a relaxed Forward-Filtering Backward-Sampling (FFBS) approach.
As a reparameterized gradient estimator, the Gumbel-CRF gives more stable gradients than score-function based estimators. 
As a structured inference network, we show that it learns interpretable templates during training, 
which allows us to control the decoder during testing. 
We demonstrate the effectiveness of our methods with experiments on data-to-text generation and unsupervised paraphrase generation. 
\end{abstract}
 
\section{Introduction}
\label{sec:introduction}

Recent work in NLP has focused on model \textit{interpretability} and \textit{controllability}~\citep{wiseman2018learning,Li2020PosteriorCO,hu2017toward,shen2017style,Fu2019ParaphraseGW}, 
aiming to add transparency to black-box neural networks and control model outputs with task-specific constraints. 
For tasks such as data-to-text generation~\citep{puzikov2018e2e,wiseman2018learning} or paraphrasing~\citep{Liu2019UnsupervisedPB, Fu2019ParaphraseGW}, 
interpretability and controllability are especially important as users are interested in what linguistic properties 
-- e.g., syntax~\citep{bao-etal-2019-generating}, phrases~\citep{wiseman2018learning}, main entities~\citep{Puduppully2019DatatotextGW} and lexical choices~\citep{Fu2019ParaphraseGW} -- 
are controlled by the model
and which part of the model controls the corresponding outputs.  

Most existing work in this area relies on non-probabilistic approaches or on complex Reinforcement Learning (RL)-based hard structures. 
Non-probabilistic approaches include using attention weights as sources of interpretability~\citep{Jain2019AttentionIN, Wiegreffe2019AttentionIN}, 
or building specialized network architectures like entity modeling~\citep{Puduppully2019DatatotextGW} or copy mechanism~\citep{gu2016incorporating}.
These approaches take advantages of differentiability and end-to-end training,
but does not incorporate the expressiveness and flexibility of probabilistic approaches~\citep{murphy2012machine, Blei2016VariationalIA,kingma2013auto}. 
On the other hand, approaches using probabilistic graphical models usually involve non-differentiable sampling~\citep{kim2019unsupervised,yin2018structvae,Li2020PosteriorCO}. 
Although these structures exhibit better interpretability and controllability~\citep{Li2020PosteriorCO}, 
it is challenging to train them in an end-to-end fashion.   

In this work, we aim to combine the advantages of relaxed training and graphical models,  focusing on conditional random field (CRF) models. 
Previous work in this area primarily utilizes the score function estimator (aka. REINFORCE)~\citep{williams1992simple, ranganath2013black,kim2019unsupervised,li2019dependency}
to obtain Monte Carlo (MC) gradient estimation for simplistic categorical models~\citep{Mohamed2019MonteCG,Mnih2016VariationalIF}.
However, given the combinatorial search space, these approaches
suffer from high variance~\citep{greensmith2004variance} and are notoriously difficult to train~\citep{kim2019unsupervised}. 
Furthermore, in a linear-chain CRF setting, score function estimators can only provide gradients for the \textit{whole} sequence, 
while it would be ideal if we can derive fine-grained pathwise gradients~\citep{Mohamed2019MonteCG} for \textit{each step} of the sequence.  
In light of this, naturally one would turn to reparameterized estimators with pathwise gradients which are known to be more stable with lower variance~\citep{kingma2013auto,Mohamed2019MonteCG}.

Our simple approach for reparameterizing CRF inference is to directly relax the sampling process itself. 
Gumbel-Softmax~\citep{jang2016categorical,maddison2016concrete} has become a popular method for relaxing categorical sampling. 
We propose to utilize this method to relax each step of CRF sampling utilizing the forward-filtering backward-sampling algorithm~\citep{murphy2012machine}. 
Just as with Gumbel-Softmax, this approach becomes exact as temperature goes to zero, and provides a soft relaxation in other cases. 
We call this approach \textit{Gumbel-CRF}.
As is discussed by previous work that a structured latent variable may have a better inductive bias for capturing the discrete nature of sentences~\citep{kim2018tutorial, kim2019unsupervised,Fu2019ParaphraseGW}, 
we apply Gumbel-CRF as the inference model in a structured variational autoencoder for learning latent templates that control the sentence structures.
Templates are defined as a sequence of states where each state controls the content (e.g., properties of the entities being discussed) of the word to be generated. 

Experiments explore the properties and applications of the Gumbel-CRF approach. 
As a \textit{reparameterized gradient estimator}, compared with score function based estimators, Gumbel-CRF not only gives lower-variance and fine-grained gradients for each sampling step,
which leads to a better text modeling performance, 
but also introduce practical advantages with significantly fewer parameters to tune and faster convergence ($\mathsection$ \ref{ssec:exp_density}). 
As a \textit{structured inference network}, like other hard models trained with REINFORCE, 
Gumbel-CRF also induces interpretable and controllable templates for generation. 
We demonstrate the interpretability and controllability  on unsupervised paraphrase generation and data-to-text generation  ($\mathsection$ \ref{ssec:generation}). 
Our code is available at \url{https://github.com/FranxYao/Gumbel-CRF}.

\section{Related Work}
\label{sec:related_works}

\textbf{Latent Variable Models and Controllable Text Generation.} 
Broadly, our model follows the line of work on deep latent variable models \citep{yao_dgm4nlp, kingma2013auto, rezende2014stochastic, Higgins2017betaVAELB, doersch2016tutorial} 
for text generation \citep{kim2018tutorial, miao2017deep, Fu2019ParaphraseGW, zhao2017adversarially}.
% \sr{Give a bit more background on these papers.}
At an intersection of graphical models and deep learning, 
these works aim to embed interpretability and controllability into neural networks with continuous~\citep{bowman2015generating,zhao2017adversarially}, 
discrete~\citep{jang2016categorical,maddison2016concrete}, or structured latent variables~\citep{kim2019unsupervised}.
One typical template model is the Hidden Semi-Markov Model (HSMM), proposed by \citet{wiseman2018learning}.  
They use a neural generative HSMM model for joint learning the latent and the sentence with exact inference.
% We use a structured variational autoencoder to separate the inference and the generative model. 
\citet{Li2020PosteriorCO} further equip a Semi-Markov CRF with posterior regularization \citep{Ganchev10a}. 
While they focus on regularizing the inference network, we focus on reparameterizing it. 
% \citet{ye2020variational} propose a variational template machine with a continuous latent variable and multiple regularization techniques for stabilizing training. 
% Compared with their model, our model is fully unsupervised and does not require the troublesome regularization.  
Other works about controllability include \citep{shen2017style, hu2017toward, li2018delete},  
but many of them stay at word-level \citep{fu2019rethinking} while we focus on structure-level controllability. 

\textbf{Monte Carlo Gradient Estimation and Continuous Relaxation of Discrete Structures.} 
Within the range of MC gradient estimation~\citep{Mohamed2019MonteCG}, 
our Gumbel-CRF is closely related to reparameterization and continuous relaxation techniques for discrete structures~
\citep{xie2019differentiable, linderman2017reparameterizing, kool2019stochastic}. 
To get the MC gradient for discrete structures, many previous works use score-function estimators \citep{ranganath2013black, Mnih2016VariationalIF, kim2019unsupervised, yin2018structvae}. 
This family of estimators is generally hard to train, especially for a structured model \citep{kim2019unsupervised}, 
while reparameterized estimators \citep{kingma2013auto,rezende2014stochastic} like Gumbel-Softmax \citep{jang2016categorical, maddison2016concrete} give a more stable gradient estimation.
% \citet{jang2016categorical, maddison2016concrete} propose the Gumbel-Softmax, or the Concrete distribution, to reparameterize categorical samples. 
In terms of continuous relaxation, the closest work is the differentiable dynamic programming proposed by \citet{mensch2018differentiable}.
% They further give a relaxed Viterbi (RViterbi) algorithm as an example instantiation. 
However, their approach takes an optimization perspective, and it is not straightforward to combine it with probabilistic models.  
Compared with their work, our Gumbel-CRF is a specific differentiable DP tailored for FFBS with Gumbel-Softmax. 
In terms of reparameterization, the closest work is the Perturb-and-MAP Markov Random Field (PM-MRF), proposed by \citet{papandreou2011perturb}.
% \citet{papandreou2011perturb} propose Perturb-and-Map Markov Random Field (PM-MRF), further applied to dependency parsing \citep{corro2018differentiable}.
However, when used for sampling from CRFs, PM-MRF is a biased sampler, while FFBS is unbiased. 
We will use a continuously relaxed PM-MRF as our baseline, and compare the gradient structures in detail in the Appendix.
% If we combine this technique with the RViterbi algorithm, we will get a reparameterized but biased sampler for CRFs while our Gumbel-FFBS gives unbiased samples 
% \sr{<- this is misleading, Gumbel-CRF is not unbiased}. 
% We will give a more detailed discussion of these works in section \ref{sec:gumbel_crf}.

% \sr{It is really important that you make clear the differences between your methods and these approaches. Do it high-level here, but then do it formally in section 3}
% There is a similar technique named 
% But this method gives a biased sampler while ours is unbiased. 

% Other works include continuous relaxations for other structures like:  
% sets, 
% permutation\citep{linderman2017reparameterizing, mena2018learning}, and beam search \citep{kool2019stochastic}. 

% \textbf{Unsupervised Structured Prediction.} 
% Our work is also related to unsupervised structured prediction works like the Unsupervised Recurrent Neural Network Grammars (URNNGs) \citep{kim2019unsupervised}. 
% URNNGs use TreeCRFs\citep{eisner2000bilexical} in the middle, while we use linear-chain. 
% Since TreeCRFs are not reparameterizable, they use the score function estimator\citep{ranganath2013black}, while we develop a Gumbel-CRF reparameterization. 
% Other related works include CRF-Autoencoders \citep{ammar2014conditional}, Struct-VAE \citep{yin2018structvae}, and VIB \citep{li2019specializing}. 
% This line of work focus on how to infer meaningful structures, while we focus on how to use these structures for generation.

%\section{Gumbel-Softmax for Categorical Reparameterization}
%\label{sec:gumbel}
\section{Gumbel-CRF: Relaxing the FFBS Algorithm}
\label{sec:gumbel_crf}

%  \fy{Generally do not change this section, but make it clear that Argmax is differentiable. Also make it clear what kind of roles that relaxation and reparameterization is playing. Relaxation: to change discrete to continuous; Reparameterization: differentiable sampling.}

%
% \begin{algorithm}[t]
%   \centering
%   \caption{\footnotesize Entropy Calculation for Linear-chain CRFs}\label{algo:entropy}
%   \footnotesize
%   \begin{algorithmic}[1]
%     \State \text{\textbf{Input:} $\Phi(z_{t-1} = i, z_t = j, x_t), \vec{\alpha}_{1:T}$; \quad \textbf{Initialization:} $\mathcal{H}_1(i) = 0$. } 
%     \For{$t \gets 1, T - 1$}
%     \State $w_{t + 1}(i, j) = \frac{\Phi(z_t=i, z_{t + 1} = j, x_{t + 1})}{\vec{\alpha}_{t + 1}(j)}$; \quad $\mathcal{H}_{t + 1}(j) = \sum_i w_{t + 1}(i, j)[\mathcal{H}_t(i) - \log w_{t + 1}(i, j)]$
%     \EndFor
%     \State $w_T = \vec{\alpha}_T / Z$;\quad $\mathcal{H}(Y | X) = \sum_j w_T(j)[\mathcal{H}_T(j) - w_T(j)]$
%     \State \textbf{Return}  $\mathcal{H}(Y | X)$ 
%   \end{algorithmic}
% \end{algorithm}
%
%
\begin{figure}[t]
\begin{minipage}{0.48\textwidth}
  \begin{algorithm}[H]
      \centering
      \caption{\footnotesize Forward Filtering Backward Sampling}\label{algo:FFBS}
      \footnotesize
      \begin{algorithmic}[1]
          \State \text{\textbf{Input:} $\Phi(z_{t-1}, z_t, x_t), t \in\{1, .., T\}, \alpha_{1:T}, Z$} 
          \State \text{Calculate $p(z_T | x) = \alpha_T / Z$}
          \State \text{Sample $\hat{z}_T \sim p(z_T | {x})$}
          \For{$t \gets T - 1, 1$}
          \State $p(z_{t} | \hat{z}_{t + 1}, {x}) = \frac{\Phi(z_{t}, \hat{z}_{t + 1}, x_{t + 1})\alpha_{t}(z_{t})}{\alpha_{t + 1}(\hat{z}_{t + 1})}$ 
          \State Sample $\hat{z}_t \sim p(z_{t} | \hat{z}_{t + 1}, {x})$
          \EndFor
          \State \textbf{Return}  $\hat{z}_{1:T}$ %\Comment{$\hat{z}$ is not differentiable}
      \end{algorithmic}
  \end{algorithm} 
  \includegraphics[width=\textwidth]{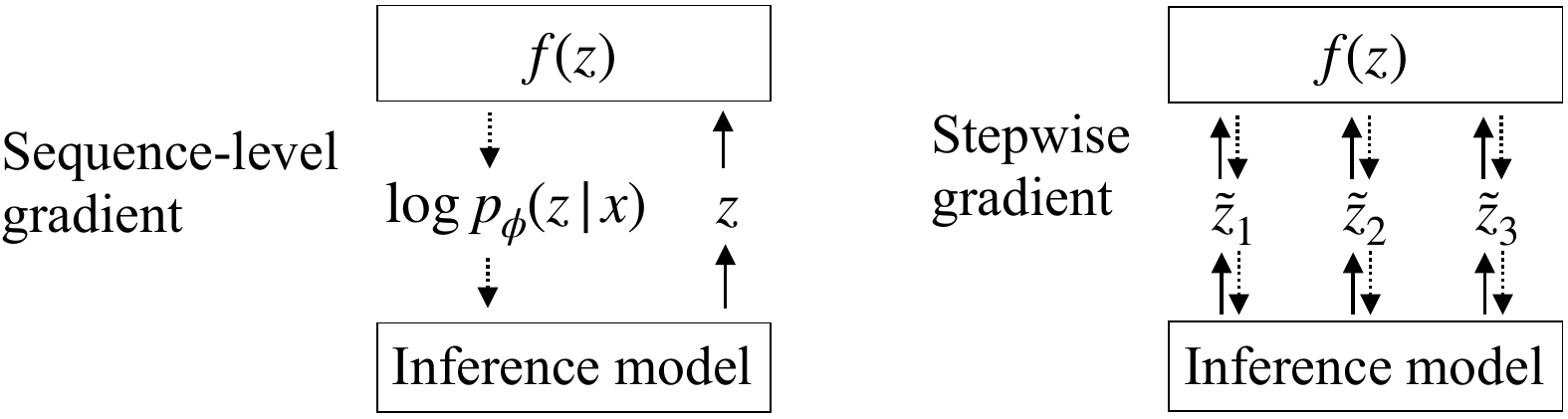} 
  %   \caption{\sr{This picture is (still) sloppy, make all the fonts the same. I also don't get it or understand the vector.}}
  \end{minipage}
  \hfill
  \begin{minipage}{0.48\textwidth}
    \begin{algorithm}[H]
      \centering
      \caption{\footnotesize Gumbel-CRF (Forward Filtering Backward Sampling with Gumbel-Softmax)}\label{algo:FFBSGS}
      \footnotesize
      \begin{algorithmic}[1]
          \State \text{\textbf{Input:} $\Phi(z_{t-1}, z_t, x_t), t \in\{1, .., T\}, \alpha_{1:T}, Z$} 
          \State \text{Calculate:}
          \State \text{\;\;\;$\pi_T = \alpha_T / Z$}
          \State \text{\;\;\;$\tilde{z}_T =$ softmax$((\log \pi_T + g) / \tau)$, $g \sim \text{G}(0)$}
          \State \text{\;\;\;$\hat{z}_T = \text{argmax}(\tilde{z}_T)$}
          \For{$t \gets T - 1, 1$}
          \State $\pi_t = \frac{\Phi(z_{t}, \hat{z}_{t + 1}, x_{t + 1})\alpha_{t}(z_{t})}{\alpha_{t + 1}(\hat{z}_{t + 1})}$ 
          \State \text{$\tilde{z}_t =$ softmax$((\log \pi_t + g) / \tau)$, $g \sim \text{G}(0)$}
          \State \text{$\hat{z}_t = \text{argmax}(\tilde{z}_t)$}
          \EndFor
          \State \textbf{Return}  $\hat{z}_{1:T}, \tilde{z}_{1:T}$ \Comment{$\tilde{z}$ is a relaxation for $\hat{z}$}  
      \end{algorithmic}
  \end{algorithm}
  \end{minipage}
  \caption{\label{fig:gumbel_crf} \small 
  Gumbel-CRF FFBS algorithm and visualization for sequence-level v.s. stepwise gradients. Solid arrows show the forward pass and dashed arrows show the backward pass. 
  }
\end{figure}

In this section, we discuss how to relax a CRF with Gumbel-Softmax to allow for reparameterization.
In particular, we are interested in optimizing $\phi$ for an expectation under a parameterized CRF distribution, e.g.,
\begin{equation}
 \mathbb{E}_{p_\phi(z | x)}[f(z)] 
\end{equation}

We start by reviewing Gumbel-Max~\citep{jang2016categorical}, a technique for sampling from a categorical distribution. 
Let $Y$ be a categorical random variable with domain $\{1, .., K\}$ parameterized by the probability vector $\mathbf{\pi} = [\pi_1, .., \pi_K]$, 
denoted as $y \sim \text{Cat}(\mathbf{\pi})$. 
Let $\text{G}(0)$ denotes the standard Gumbel distribution, $g_i \sim \text{G}(0), i \in \{1, .., K\}$ are i.i.d. gumbel noise. 
Gumbel-Max sampling of $Y$ can be performed as: 
  $y = \arg\max_i  (\log \pi_i + g_i)$.
Then the Gumbel-Softmax reparameterization is a continuous relaxation of Gumbel-Max by replacing the 
% non-differentiable \sr{Argmax is differentiable, use a different reasoning} 
hard argmax operation with the softmax,
\begin{equation}
  \tilde{y} = \text{softmax}([\log \pi_1 + g_1, .., \log \pi_K + g_K ]) = \frac{\exp((\log \pi + g) / \tau)}{\sum_i \exp((\log \pi_i + g_i) / \tau)}
\end{equation}
where $\tilde{y}$ can be viewed as a relaxed one-hot vector of $y$. 
% As a continous random variable, $\tilde{y}$ comes from the Gumbel-Softmax distribution. 
As $\tau \to 0$, we have $\tilde{y} \to y$. 

% \sr{This section is nice and clear. Well done. Just check for typos and small things.}

Now we turn our focus to CRFs which generalize softmax to combinatorial structures.
Given a sequence of inputs $x = [x_1, .., x_T]$ a linear-chain CRF is parameterized by the factorized potential function $\Phi(z, x)$ 
and defines the probability of a state sequence $z = [z_1, .., z_T], z_t \in \{1, 2, ..., K\}$ over $x$. 
%We focus on the case where $x$ is a sentence and $z$ is a latent template for $x$.
%
\begin{align}
  &p(z | x) = \frac{\Phi(z, x)}{Z}\quad\quad \Phi(z, x) = \prod_t \Phi(z_{t-1}, z_t, x_t) \label{eq:gibbs}\\
  &\alpha_{1:T} = \text{Forward}(\Phi(z, x))\quad\quad Z = \sum_i \alpha_T(i) \label{eq:forward}
\end{align}
The conditional probability of $z$ is given by the Gibbs distribution with a factorized potential (equation~\ref{eq:gibbs}).
The partition function $Z$ can be calculated with the Forward algorithm~\citep{sutton2012introduction} (equation~\ref{eq:forward})
where $\alpha_t$ is the forward variable summarizing the potentials up to step $t$. 
% \sr{why do you use $q$ and not $p$? I feel like better to derive this approach with $p$ and use $q$ only for the VAE. }

To sample from a linear-chain CRF, the standard approach is to use the forward-filtering backward-sampling (FFBS) algorithm~(Algorithm~\ref{algo:FFBS}). 
This algorithm takes $\alpha$ and $Z$ as inputs and samples $z$ with a backward pass utilizing the locally conditional independence. 
The hard operation comes from the backward sampling operation for $\hat{z}_t \sim p(z_t | \hat{z}_{t + 1}, x) $  at each step (line 6). This is the operation that we focus on. 

We observe that $p(z_t | \hat{z}_{t + 1}, x)$ is a categorical distribution, which can be directly relaxed with Gumbel-Softmax.
This leads to our derivation of the Gumbelized-FFBS algorithm(Algorithm~\ref{algo:FFBSGS}). 
The backbone of Algorithm \ref{algo:FFBSGS} is the same as the original FFBS except for two key modifications: 
(1) Gumbel-Max (line 8-9) recovers the \textit{unbiased sample} $\hat{z}_t$ and the same sampling path as Algorithm~\ref{algo:FFBS};
(2) the continuous relaxation of argmax with softmax (line 8) that gives the differentiable\footnote{
Note that argmax is also differentiable almost everywhere, however its gradient is almost 0 everywhere and not well-defined at the jumping point~\citep{paulus2020gradient}. 
% which further invalidates the change of expectation and differentiation \citep{paulus2020gradient}. 
Our relaxed $\tilde{z}$ does not have these problems. 
} 
(but biased)~$\tilde{z}$.

We can apply this approach in any setting requiring structured sampling.
For instance let $p_\phi(z|x)$ denote the sampled distribution  
and $f(z)$ be a downstream model.
% ($\phi$ is the model parameter) 
%and $p(x|z)$ denote the generative model, 
We achieve a reparameterized gradient estimator for the sampled model with $\tilde{z}$:
\begin{equation}
  \nabla_\phi \mathbb{E}_{p_\phi(z | x)}[f(z)] \approx \mathbb{E}_{g \sim \text{G}(0)} [\nabla_\phi f(\tilde{z}(\phi, g))]
\end{equation}
We further consider a straight-through (ST) version~\citep{jang2016categorical} of this estimator where we use the hard sample $\hat{z}$ in the forward pass, 
and back-propagate through each of the soft $\tilde{z}_t$.

We highlight one additional advantage of this reparameterized estimator (and its ST version), compared with the score-function estimator. Gumbel-CRF uses $\tilde{z}$ which recieve direct \textit{fine-grained gradients} for each step from the $f$ itself. 
As illustrated in Figure~\ref{fig:gumbel_crf} (here $f$ is the generative model), 
 $\tilde{z}_t$ is a differentiable function of the potential and the noise: $\tilde{z}_t = \tilde{z}_t(\phi, g)$.
So during back-propagation, we can take \textit{stepwise gradients} $\nabla_\phi \tilde{z}_t(\phi, g)$ weighted by the uses of the state.
On the other hand, with a score-function estimator, we only observe the reward for the whole sequence, 
so the gradient is at the \textit{sequence level}  $\nabla_\phi \log p_\phi(z|x)$. 
The lack of intermediate reward, i.e., stepwise gradients, is an essential challenge in reinforcement learning~\citep{silver2016mastering, Yu2017SeqGANSG}. While we could derive a model specific factorization 
for the score function estimator \cite{}, 
% TODO: YAO cite some of the RL for translation papers
% which further induces many modeling and engineering challenges.  
this challenge is circumvented with the structure of Gumbel-CRF, thus significantly reducing modeling complexity in practice (detailed demonstrations in experiments).
% We will give a detailed explanation of the practical advantages of substituting REINFORCE with Gumbel-CRF in the experiments. 

% \input{041gumbel_alternatives}

\section{Gumbel-CRF VAE}
\label{sec:model}

% In this section, we introduce our structured model for learning latent templates (Figure \ref{fig:model}). 

% \subsection{Generative Model}
% \label{ssec:generative_model}

An appealing use of reparameterizable CRF models is to learn variational autoencoders (VAEs) with a structured inference network. 
Past work has shown that these models (trained with RL) \citep{kim2019unsupervised,Li2020PosteriorCO} can learn to induce useful latent structure while producing accurate models. 
We introduce a VAE for learning latent templates for text generation. This model uses a fully autoregressive generative model with latent control states. To train these control states, it uses a CRF variational posterior as an inference model. Gumbel CRF is used to reduce the variance of this procedure.

\textbf{Generative Model}$\quad$
We assume a simple generative process where each word $x_t$ of a sentence $x = [x_1, .., x_T]$ 
is controlled by a sequence of latent states $z = [z_1, .., z_T]$, i.e., template, similar to~\citet{Li2020PosteriorCO}:
\begin{align}
  p_\theta(x, z) &= \prod_t p(x_t | z_t, z_{1:t-1}, x_{1: t-1}) \cdot p(z_t | z_{1:t-1}, x_{1: t-1}) \\ 
  h_t &= \text{Dec}([z_{t-1}; x_{t-1}], h_{t -1}) \\
  p(z_t | z_{1:t-1}, x_{1: t-1}) &= \text{softmax}(\text{FF}(h_t)) \\ 
  p(x_t | z_t, z_{1:t-1}, x_{1: t-1}) &= \text{softmax}(\text{FF}([e(z_t); h_t]))
\end{align}
Where Dec$(\cdot)$ denotes the decoder, 
FF$(\cdot)$ denotes a feed-forward network, 
$h_t$ denotes the decoder state, 
$[\cdot; \cdot]$ denotes vector concatenation.
$e(\cdot)$ denotes the embedding function. 
Under this formulation, the generative model is autoregressive w.r.t. both $x$ and $z$.
Intuitively, it generates the control states and words in turn, and the current word $x_t$ is primarily controlled by its corresponding $z_t$.

\begin{figure}[t]
  \centering
  \includegraphics[width=\textwidth]{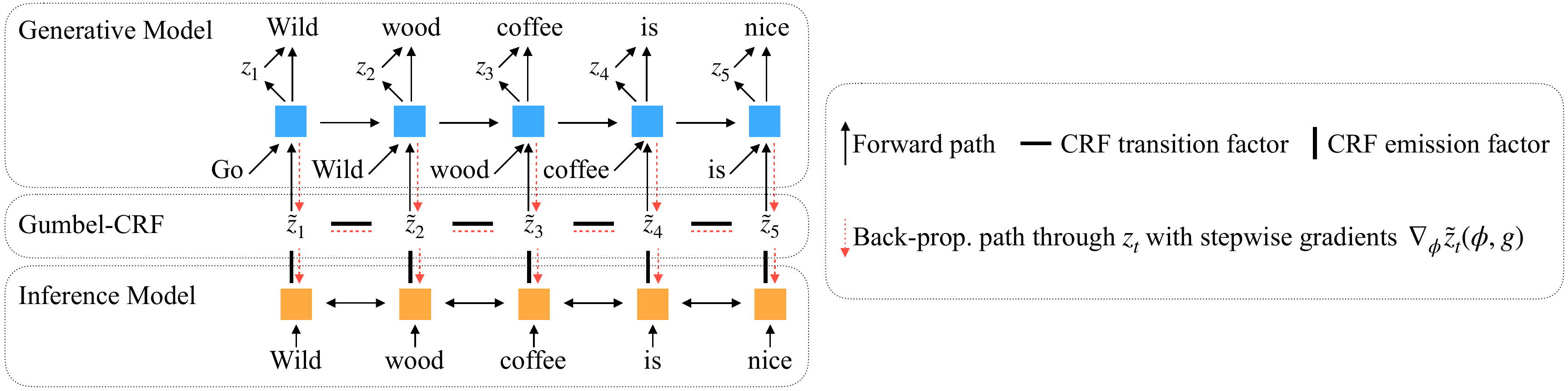}  
  \caption{\label{fig:model} \small 
  Architecture of our model.
  Note the structure of gradients induced by Gumbel-CRF differs significantly from score-function approaches.
  Score function receives a gradient for the sampled sequence $\nabla_\phi \log q_\phi(z | x)$ 
  while Gumbel-CRF allows the model to backprop gradients along each sample step $\nabla_\phi \tilde{z}_t$ (red dashed arrows) without explicit factorization of the generative model. 
  }
\end{figure}  

% \subsection{Inference Model}
\textbf{Inference Model}$\quad$
Since the exact inference of the posterior $p_\theta(z | x)$ is intractable, we approximate it with a variational posterior $q_\phi(z | x)$ and
optimize the following form of the ELBO objective:
\begin{equation}
  \text{ELBO} = \mathbb{E}_{q_\phi(z | x)}[\log p_\theta(x, z)] - \mathcal{H}[q_\phi(z | x)] \label{eq:elbo}
\end{equation}
Where $\mathcal{H}$ denotes the entropy. 
The key use of Gumbel-CRF is for reparameterizing the inference model $q_\phi({z} | {x})$ to learn control-state templates. 
Following past work~\citep{huang2015bidirectional}, we parameterize $q_\phi({z} | {x})$ as a linear-chain CRF whose potential is predicted by a neural encoder: 
\begin{align}
  h^{(\text{enc})}_{1:T} &= \text{Enc}(x_{1:T}) \\
  \Phi(z_t, x_t) &= W_\Phi h^{(\text{enc})}_t + b_\Phi \\ 
  \Phi(z_{t-1}, z_t, x_t) &= \Phi(z_{t-1}, z_t) \cdot  \Phi(z_t, x_t)
\end{align}
Where Enc$(\cdot)$ denotes the encoder and $h^{(\text{enc})}_t$ is the encoder state. 
With this formulation, the entropy term in equation \ref{eq:elbo} can be computed efficiently with dynamic programming, which is differentiable~\citep{mann2007efficient}.

\textbf{Training and Testing}$\quad$
% \sr{put all the training details here -> } 
The key challenge for training is to maximize the first term of the ELBO under the expectation of the inference model, i.e.
\[\mathbb{E}_{q_\phi(z | x)}[\log p_\theta(x, z)] \]
Here we use the Gumbel-CRF gradient estimator with relaxed samples $\tilde{z}$ from the Gumbelized FFBS (Algorithm \ref{algo:FFBSGS}) in both the forward and backward passes.
For the ST version of Gumbel-CRF, we use the exact sample $\hat{z}$ in the forward pass and back-propogate gradients through the relaxed $\tilde{z}$.
During testing, for evaluating paraphrasing and data-to-text, we use greedy decoding for both $z$ and $x$. 
For experiments controlling the structure of generated sentences, 
we sample a fixed MAP $z$ from the training set (i.e., the aggregated variational posterior), 
feed it to each decoder steps, and use it to control the generated $x$.

\textbf{Extension to Conditional Settings}$\quad$ For conditional applications, such as paraphrasing and data-to-text, we make a conditional extension where the generative model is conditioned on a source data structure $s$, formulated as $p_\theta(x, z | s)$. 
Specifically, for paraphrase generation, $s = [s_1, ..., s_N]$ is the bag of words (a set, $N$ being the size of the BOW) of the source sentence, similar to~\citet{Fu2019ParaphraseGW}.
We aim to generate a different sentence $x$ with the same meaning as the input sentence. 
In addition to being autoregressive on $x$ and $z$, the decoder also attend to~\citep{bahdanau2015neural} and copy~\citep{gu2016incorporating} from $s$. 
For data-to-text, we denote the source data is formed as a table of key-value pairs: $s = [(k_1, v_1), ..., (k_N, v_N)]$, N being size of the table. 
We aim to generate a sentence $x$ that best describes the table. 
Again, we condition the generative model on $s$ by attending to and copying from it. 
Note our formulation would effectively become \textit{a neural version of slot-filling}: for paraphrase generation we fill the BOW into the neural templates, and for data-to-text we fill the values into neural templates. 
We assume the inference model is independent from the source $s$ and keep it unchanged, i.e., $q_\phi(z | x, s) = q_\phi(z | x)$. 
The ELBO objective in this conditional setting is:
\begin{equation}
  \text{ELBO} = \mathbb{E}_{q_\phi(z | x)}[\log p_\theta(x, z | s)] - \mathcal{H}[q_\phi(z | x)]
\end{equation} 

\section{Experimental Setup} 
\label{sec:exp_setup}

Our experiments are in two parts. 
First,  we compare Gumbel-CRF to other common gradient estimators on the standard text modeling task. 
Then we integrate Gumbel-CRF to real-world models, specifically paraphrase generation and data-to-text generation.

%We show that like the hard models trained with REINFORCE \citep{Li2020PosteriorCO}, models trained with Gumbel-CRF also induce meaningful templates for interpretability and controllability with similar or better end-task performance (e.g., BLEU). 
%Furthermore, we show that the simplexity of Gumbel-CRF gives many practical benefits including fewer parameters to tune and faster training time. 

\textbf{Datasets}$\quad$
We focus on two datasets. 
For text modeling and data-to-text generation, we use the \texttt{E2E} dataset\citep{puzikov2018e2e}, 
a common dataset for learning structured templates for text \citep{wiseman2018learning,Li2020PosteriorCO}.
This dataset contains approximately 42K training, 
4.6K validation and 4.6K testing sentences. 
The vocabulary size is 945. 
For paraphrase generation we follow the same setting as \citet{Fu2019ParaphraseGW}, and use the common \texttt{MSCOCO} dataset. 
This dataset has 94K training and 23K testing instances. 
The vocabulary size is 8K.

\textbf{Metrics}$\quad$
For evaluating the gradient estimator performance, 
we follow the common practice and primarily compare the test negative log-likelihood (NLL) estimated with importance sampling. 
% We use 100 importance samples, which we find being enough to obtain a low-variance estimate of NLL. 
We also report relative metrics: ELBO, perplexity (PPL), and entropy of the inference network. 
Importantly, to make all estimates unbiased, all models are evaluated in \textit{a discrete setting with unbiased hard samples}.
For paraphrase task performance, we follow \citet{Liu2019UnsupervisedPB, Fu2019ParaphraseGW} and use BLEU (bigram to 4-gram)~\citep{papineni2002bleu} and ROUGE~\citep{lin2002manual} (R1, R2 and RL) to measure the generation quality.
We note that although being widely used, the two metrics do not penalize the similarity between the generated sentence and the input sentence (because we do not want the model to simply copy the input).
So we adopt iBLUE~\citep{sun-zhou-2012-joint}, a specialized BLUE score that penalize the ngram overlap between the generated sentence and the input sentence, and use is as our primary metrics. 
The iBLUE score is defined as: iB$(i, o, r) = \alpha \text{B}(o, r) + (1 - \alpha) \text{B}(o, i)$, where iB$(\cdot)$ denotes iBLUE score, B$(\cdot)$ denotes BLUE score, $i, o, r$ denote input, output, and reference sentences respectively. 
We follow \citet{Liu2019UnsupervisedPB} and set $\alpha=0.9$. 
For text generation performance, we follow \citet{Li2020PosteriorCO} and 
use the E2E official evaluation script,
which measures BLEU, NIST~\citep{belz2006comparing}, ROUGE, 
CIDEr~\citep{vedantam2015cider}, and METEOR~\citep{banerjee2005meteor}. 
More experimental details are in the Appendix. 

\paragraph{VAE Training Details} At the beginning of training, to prevent the decoder from ignoring $z$, we apply word dropout~\citep{bowman2015generating}, 
i.e., to randomly set the input word embedding at certain steps to be 0.
After $z$ converges to a meaningful local optimal, we gradually decrease word dropout ratio to 0 and recover the full model. 
For optimization, we add a $\beta$ coefficient to the entropy term, as is in the $\beta$-VAE~\citep{Higgins2017betaVAELB}. 
As is in many VAE works~\citep{bowman2015generating, dieng2018avoiding}, we observe the posterior collapse problem where $q(z|x)$ converges to meaningless local optimal. 
We observe two types of collapsed posterior in our case: 
a constant posterior ($q$ outputs a fixed $z$ no matter what $x$ is. This happens when $\beta$ is too weak), and a uniform posterior (when $\beta$ is too strong). 
To prevent posterior collapse, $\beta$ should be carefully tuned to achieve a balance.

\section{Results}  
\label{sec:experimental_setup}

\begin{table*}[t]
  \small
  \caption{\label{tab:density} \small 
  Density Estimation Results. 
  NLL is estimated with 100 importance samples. 
  Models are selected from 3 different random seeds based on validation NLL.
  All metrics are evaluated on the discrete (exact) model.  
  }
  \begin{center}
  \begin{tabular}{@{}lccccc@{}}   

  % \multicolumn{8}{c}{\texttt{Test}} \\ 
  \toprule
  % Model &  Bleu & Nist & Rouge & Cider & Meteor & I-Bleu &  I-Rouge \\ 
  %  & & \multicolumn{4}{c}{Dev} & \multicolumn{4}{c}{Test} \\ 
  Model                 & Neg. ELBO & NLL   & PPL   & Ent. & \#sample\\ 
  \hline
  RNNLM                 & -         & 34.69 & 4.94  & - & - \\
  PM-MRF                & 69.15     & 50.22 & 10.41 & 4.11 & 1 \\ 
  PM-MRF-ST             & 53.16     & 37.03 & 5.48  & 2.04 & 1\\ 
  REINFORCE-MS          & 35.11     & 34.50 & 4.84  & 3.48 & 5 \\ 
  REINFORCE-MS-C        & 34.35     & 33.82 & 4.71  & 3.34 & 5\\
  Gumbel-CRF (ours)     & 38.00     & 35.41 & 4.71  & 3.03 & 1\\ 
  Gumbel-CRF-ST (ours)  & 34.18     & 33.13 & 4.54  & 3.26 & 1\\
  \bottomrule
  \end{tabular}
  \end{center}
\end{table*}

\begin{figure}[t]
  \centering
  \includegraphics[width=\textwidth]{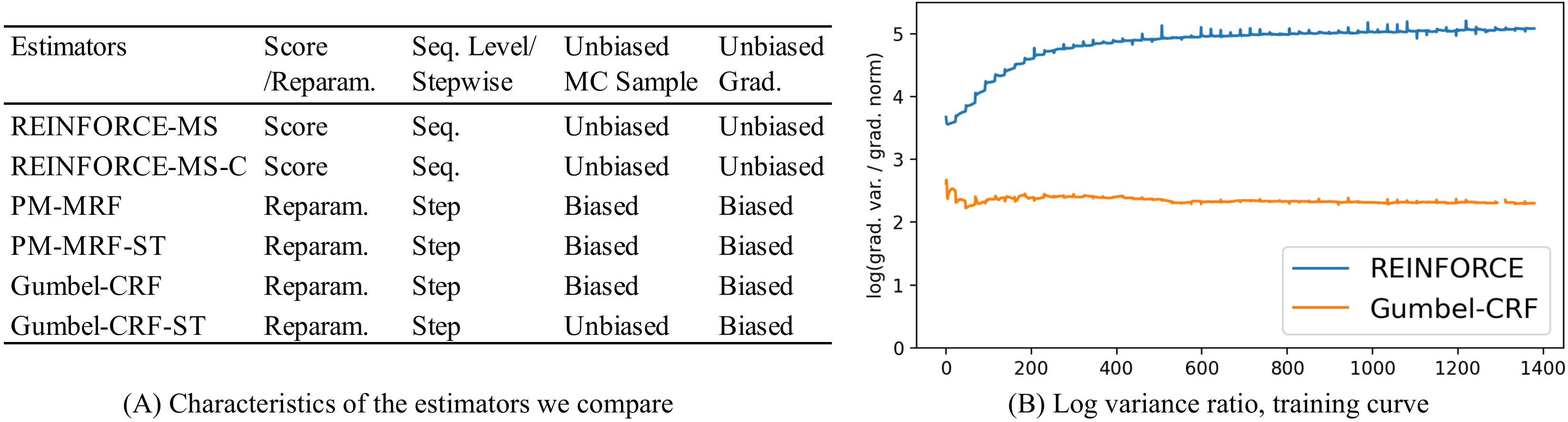}
  \caption{\label{fig:density} \small
  Text Modeling. 
  (A). Characteristics of gradient estimators.
  (B). Variance comparison, Gumbel-CRF vs REINFORCE-MS, training curve.
  % \fy{update Characteristics, update log variance, find out the stepwise version of REINFOCE for current genertive model.}
  }
\end{figure}

\subsection{Gumbel-CRF as Gradient Estimator: Text Modeling}
\label{ssec:exp_density}
We compare our Gumbel-CRF (original and ST variant) with two sets of gradient estimators: score function based and reparameterized.
For score function estimators, we compare our model with REINFORCE using the mean reward of other samples (MS) as baseline. % as is used in ~\citet{kim2019unsupervised,Li2020PosteriorCO}. 
We further find that adding a carefully tuned constant baseline helps with the scale of the gradient (REINFORCE MS-C). 
For reparameterized estimators, we use a tailored Perturb-and-Map Markov Random Field (PM-MRF) estimator~\citep{papandreou2011perturb} with the continuous relaxation introduced in \citet{corro2018differentiable}.
Compared to our Gumbel-CRF, PM-MRF adds Gumbel noise to local potentials, then runs a relaxed structured argmax algorithm~\citep{mensch2018differentiable}. 
We further consider a straight-through (ST) version of PM-MRF. 
% We note that the derivation of PM-MRF also includes complicated reparameterization and relaxation techniques. 
The basics of these estimators can be characterized from four dimensions, as listed in Figure~\ref{fig:density}(A). 
The appendix provides a further theoretical comparison of gradient structures between these estimators. 

Table~\ref{tab:density} shows the main results comparing different gradient estimators on text modeling. 
Our Gumbel-CRF-ST outperforms other estimators in terms of NLL and PPL.
% Also we see all reparameterized estimators perform better than score-function based estimators, showing the advantage of pathwise gradients. 
With fewer samples required, reparameterization and continuous relaxation used in Gumbel-CRF are particularly effective for learning structured inference networks. 
We also see that PM-MRF estimators perform worse than other estimators.
Due to the biased nature of the Perturb-and-MAP sampler, during optimization, 
PM-MRF is not optimizing the actual model. 
% PM-MRF cannot explore the full posterior space with the MC samples.
As a Monte Carlo sampler (forward pass, rather than a gradient estimator) Gumbel-CRF is less biased than PM-MRF.
We further observe that both the ST version of Gumbel-CRF and PM-MRF perform better than the non-ST version. 
We posit that this is because of the consistency of using hard samples in both training and testing time (although non-ST has other advantages).
% forward (with hard samples) and backward pass (with soft samples) in ST estimators  
% Lastly, we highlight the performance of the REINFORCE-Stepwise estimator.
% When equipped with stepwise rewards (thus similar to the reparameterized ones), it outperforms other score-function estimators with sequence-level rewards. 
% This supports our claim that an advantage of reparameterized estimators is the stepwise gradient. 

\textbf{Variance Analysis}$\quad$
% \fy{TODO: upate variance.}
To show that reparameterized estimators have lower variance, 
we compare the log variance ratio of Gumbel-CRF and REINFORCE-MS-C (Figure~\ref{fig:density}~B),
which is defined as $r = \log(\text{Var}(\nabla_\phi \mathcal{L}) / |\nabla_\phi \mathcal{L}|)$ ($\nabla_\phi \mathcal{L}$ is gradients of the inference model)\footnote{
Previous works compare variance, rather than variance ratio~\citep{Tucker2017REBARLU, Grathwohl2018BackpropagationTT}. 
We think that simply comparing the variance is only reasonable when the scale of gradients are approximately the same, which may not hold in different estimators. 
In our experiments, we observe that the gradient scale of Gumbel-CRF is significantly smaller than REINFORCE, thus the variance ratio may be a better proxy for measuring stability.  
}.
% The higher $r$ is, the higher the variance is. 
% Figure \ref{fig:density}~(B) shows that generally, reparameterized estimators have lower variance than score-function estimators. 
We see that Gumbel-CRF has a lower training curve of $r$ than REINFORCE-MS, showing that it is more stable for training. 
% Reparameterized estimators with lower variance and finer-grained gradients give better performance.  

% \paragraph{Controllable Generation}
\subsection{Gumbel-CRF for Control: Data-to-Text and Paraphrase Generation}
\label{ssec:generation}

\textbf{Data-to-Text Generation}$\quad$
Data-to-text generation models generate descriptions for tabular information.
Classical approaches use rule-based templates with better interpretability, while recent approaches use neural models for better performance. 
Here we aim to study the interpretability and controllability of latent templates.
We compare our model with neural and template-based models. 
Neural models include: D\&J\citep{duvsek2016sequence} (with basic seq2seq);
and KV2Seq\citep{fu2018natural} (with SOTA neural memory architectures);
Template models include:
SUB\citep{duvsek2016sequence} (with rule-based templates); 
hidden semi-markov model (HSMM)\citep{wiseman2018learning} (with neural templates);
and another semi-markov CRF model (SM-CRF-PC)\citep{Li2020PosteriorCO} (with neural templates and posterior regularization\citep{Ganchev10a}). 

% More details are in the appendix. 
Table \ref{tab:table_to_text} shows the results of data-to-text generation.
As expected, neural models like KV2Seq with advanced architectures achieve the best performance.
Template-related models all come with a certain level of performance costs for better controllability.  
% Our model performs similarly to the HSMM model. 
% \sr{This is true, but it doesn't tell me what I should take away from this section. What does this mean? What does it imply about your research questions?}
Among the template-related models, SM-CRF PC performs best. 
However, it utilizes multiple weak supervision to achieve better template-data alignment, 
while our model is fully unsupervised for the template.
Our model, either trained with REINFORCE or Gumbel-CRF, outperforms the baseline HSMM model. 
We further see that in this case, models trained with Gumbel-CRF gives better end performance than REINFORCE.

% We emphasize that our approach induces interpretability and controllability, as explained below. 

\begin{table*}[t!]
  \small
  \caption{\label{tab:table_to_text} \small Data-to-text generation results. 
  Upper: neural models, Lower: template-related models. Models are selected from 5 different random seeds based on validation BLEU. 
  }
  \begin{center}
  \begin{tabular}{@{}lccccc@{}}   
  \toprule
  Model &  BLEU & NIST & ROUGE & CIDEr & METEOR\\ 
  \hline
  D\&J\citep{duvsek2016sequence}        & 65.93 & 8.59 & 68.50 & 2.23 & 44.83 \\ 
  KV2Seq\citep{fu2018natural}    & 74.72 & 9.30 & 70.69 & 2.23 & 46.15  \\ \hline  
  % KV2Seq-sampling & 67.79 & 8.85 & 67.42 & 2.10 & 43.57 \\ \hline
  % KV2S-sampling aggresive & 51.80 & 7.70 & 58.65 & 1.56 & 39.02 & \bf 30.81 & \bf 39.21 \\ \hline 
  SUB\citep{duvsek2016sequence}         & 43.78 & 6.88 & 54.64 & 1.39 & 37.35 \\ 
  HSMM\citep{wiseman2018learning}       & 55.17 & 7.14 & 65.70 & 1.70 & 41.91 \\
  HSMM-AR\citep{wiseman2018learning}    & 59.80 & 7.56 & 65.01 & 1.95 & 38.75 \\
  SM-CRF PC \citep{Li2020PosteriorCO}   & 67.12 & 8.52 & 68.70 & 2.24 & 45.40 \\
  REINFORCE             & 60.41 & 7.99 & 62.54 & 1.78 & 38.04 \\
  Gumbel-CRF            & 65.83 & 8.43 & 65.06 & 1.98 & 41.44 \\
  \bottomrule
  \end{tabular}
  \end{center}
\end{table*}

\begin{figure}[t!]
  \centering
  \includegraphics[width=\textwidth]{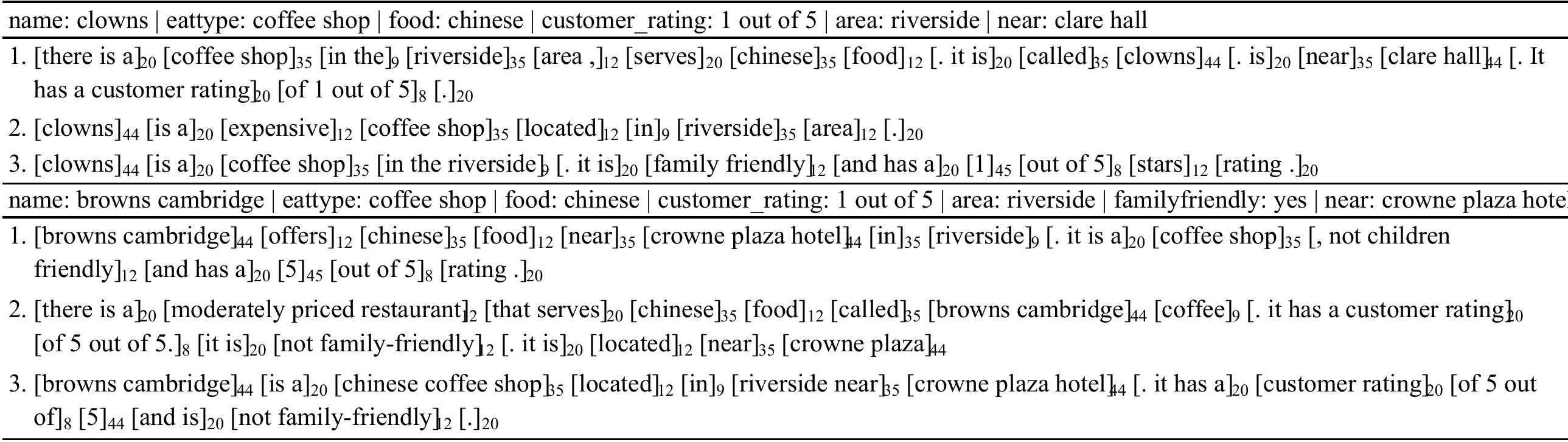}

  \caption{\label{fig:template_discussion} \small 
  Controllable generation with templates.
  }
\end{figure}

\begin{figure}[t!]
  \centering
  \includegraphics[width=\textwidth]{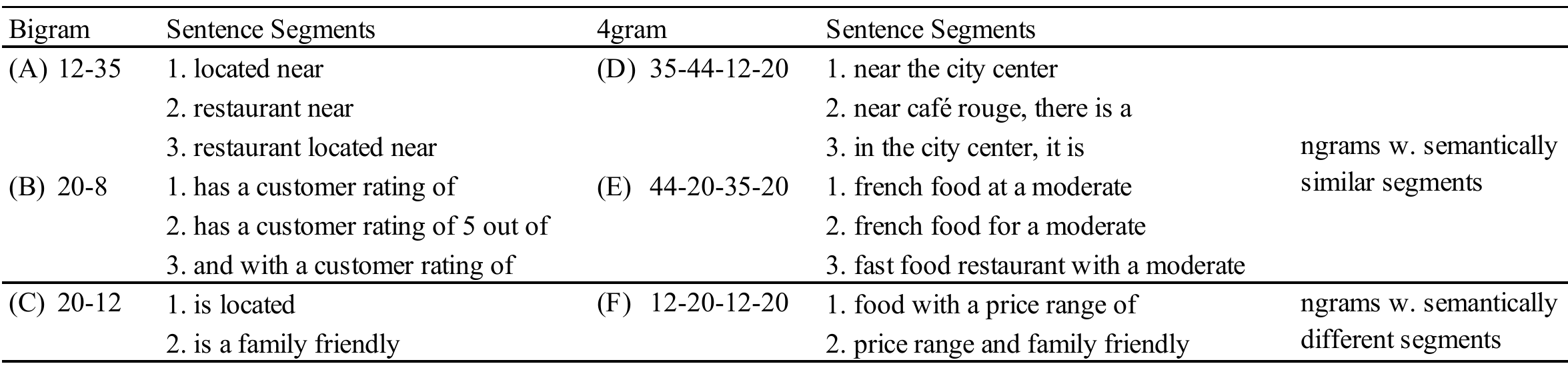}

  \caption{\label{fig:interpretability} \small 
  Analysis of state ngrams. 
  State ngrams correlate to sentence meaning. 
  In cases (A, B, D, E), semantically similar sentence segments are clustered to the same state ngrams: (A) ``location'' (B) ``rating'' (D) ``location'' (E) ``food`` and ``price''. 
  Yet there are also cases where state ngrams correspond to sentence segments with different meaning: (C1) ``location'' v.s. (C2) ``comments''; (F1) ``price'' v.s. (F2) ``price'' and ``comments''. 
  }
\end{figure} 

To see how the learned templates induce controllability, we conduct a qualitative study. 
To use the templates, after convergence, we collect and store all the MAP $z$ for the training sentences. 
During testing, given an input table, we retrieve a template $z$
%  (recall that $z$ is an expanded version of $s$ whose segment length is determined by $g$, $\mathsection$ \ref{sec:model}). 
and use this $z$ as the control state for the decoder. 
Figure~\ref{fig:template_discussion} shows sentences generated from templates. 
We can see that sentences with  different templates exhibit different structures.
E.g,. the first sentence for the \textit{clowns} coffee shop starts with the location, while the second starts with the price. 
We also observe a \textit{state-word correlation}. 
E.g,. state 44 always corresponds to the name of a restaurant and state 8 always corresponds to the rating.
% , and state 35 always corresponds to a Chinese restaurant. 

To see how learned latent states encode sentence segments, we associate frequent $z$-state ngrams with their corresponding segments (Figure~\ref{fig:interpretability}). 
Specifically, after the convergence of training, we:
(a) collect the MAP templates for all training cases, 
(b) collapse consecutive states with the same class into one single state (e.g., a state sequence [1, 1, 2, 2, 3] would be collapsed to [1, 2, 3]),
(c) gather the top 100 most frequent state ngrams and their top5 corresponding sentence segments, 
(d) pick the  mutually most different segments (because the same state ngram may correspond to very similar sentence segments, and the same sentence segment may correspond to different state ngrams).
Certain level of cherry picking happens in step (d). 
We see that state ngrams have a vague correlation with sentence meaning.
In cases (A, B, D, E), a state ngram encode semantically similar segments (e.g., all segments in case A are about location, and all segments in case E are about food and price). 
But the same state ngram may not correspond to the same sentence meaning (cases C, F). 
For example, while (C1) and (C2) both correspond to state bigram 20-12, (C1) is about location but (C2) is about comments.

% We further note that these state ngrams may not be interpreted as syntax because different syntactical phrases may be included in the same state ngram. 
% Combining this observation with the previous state-word correlation, 
% we can say that a template jointly encodes the lexical and structural information, where the lexical information is encoded by individual states, and the structural information is encoded by the sequential combination (ngrams) of states.

 \textbf{Unsupervised Paraphrase Generation} $\quad$
% As a benchmark for testing generative models for unconditional text generation~\citep{Miao2018CGMHCS, bao-etal-2019-generating},
Unsupervised paraphrase generation is defined as generating different sentences conveying the same meaning of an input sentence without parallel training instances. 
% \sr{Short description of what this task is and why it is relevant here}
To show the effectiveness of Gumbel-CRF as a gradient estimator, we compare the results when our model is trained with REINFORCE. 
To show the overall performance of our structured model, 
we compare it with other unsupervised models, including:
a Gaussian VAE for paraphrasing \citep{bowman2015generating};
CGMH \citep{Miao2018CGMHCS}, a general-purpose MCMC method for controllable generation; 
UPSA \citep{Liu2019UnsupervisedPB}, a strong paraphrasing model with simulated annealing. 
To better position our template model, we also report the supervised performance of a state-of-the-art latent bag of words model (LBOW) \citep{Fu2019ParaphraseGW}.

\begin{table*}[t!]
  \small
  \caption{\label{tab:paraphrase} \small Paraphrase Generation. Upper: supervised models, Lower: unsupervised models.  
  Models are selected from 5 random seeds based validation iB4 (iBLUE 4 gram). 
  % Compared with the unsupervised baseline Gaussian VAE, our model achieves significantly better quality and similar diversity. 
  % We observe a reasonable quality gap to the supervised model, and better quality than the unsupervised Gausian VAE.
% \sr{What does AD and NA mean? Also why do you call it Template CRF?}
  }
  \begin{center}
  \begin{tabular}{@{}lccccccc@{}} 
  \toprule
  Model &  iB4 & B2 &  B3 & B4 &  R1 &  R2 &  RL \\ 
  \hline
  LBOW \citep{Fu2019ParaphraseGW} & - & 51.14 & 35.66 & 25.27 & 42.08 & 16.13 & 38.16  \\  
  \hline
  Gaussian VAE\citep{bowman2015generating} & 7.48 & 24.90 & 13.04 & 7.29 & 22.05 & 4.64 & 26.05\\ 
  CGMH \citep{Miao2018CGMHCS} & 7.84 & - & - & 11.45 & 32.19 &  8.67 & - \\
  UPSA \citep{Liu2019UnsupervisedPB} & 9.26 & - & - & 14.16 & 37.18 &  11.21 & - \\
  REINFORCE & 11.20  & 41.29 & 26.54 & 17.10 & 32.57 & 10.20 & 34.97 \\ 
  Gumbel-CRF & 10.20 & 38.98 & 24.65 & 15.75 & 31.10 & 9.24 & 33.60 \\ 
  \bottomrule
  \end{tabular}
  \end{center}
\end{table*}

Table \ref{tab:paraphrase} shows the results. 
As expected, the supervised model LBOW performs better than all unsupervised models.  
Among unsupervised models, the best iB4 results come from our model trained with REINFORCE. 
In this task, when trained with Gumbel-CRF, our model performs worse than REINFORCE (though better than other paraphrasing models). 
We note that this inconsistency between the gradient estimation performance and the end task performance involve multiple gaps between ELBO, NLL, and BLEU.
The relationship between these metrics may be an interesting future research direction. 

% We note that it is quite interesting that our model is conceptually very simple: 
% it just view the input sentence as a bag of words, and (re-)organize them into a new sentence. 
% Yet this simple approach outperforms models involving complicated techniques like UPSA on certain metrics like iBLUE. 
% This observation may open the question about what kind of inductive bias is actually required for unsupervised paraphrasing. 
% To summarize, like REINFORCE, our Gumbel-CRF is also effective for training paraphrasing models without much lose of end-performance. 
% We will further highlight the additional practical benefits of Gumbel-CRF in later sections. 

\begin{table*}[t]
  \small
  \caption{\label{tab:practical} \small Practical benefits of using Gumbel-CRF. 
  Typically, REINFORCE has a long list of parameters to tune: $h$ entropy regularization, $b_0$ constant baseline, $b$ baseline model, $r$ reward scaling, $\#s$ number of MC sample. 
  Gumbel-CRF reduces the engineering complexity with significantly less parameters  ($h$ entropy regularization, $\tau$ temperature annealing), less samples required (thus less memory consumption), and less time consumption. 
  Models tested on Nvidia P100 with batch size 100. 
  }
  \begin{center}
  \begin{tabular}{@{}lcccc@{}}   
  \toprule
  Model       & Hyperparams. & \#s & GPU mem & Sec. per batch \\ 
  \hline
  REINFORCE   & $h, b_0, b, r, \#s$ & 5 & 1.8G & 1.42 \\ 
  Gumbel-CRF  & $h, \tau$ & 1 & 1.1G & 0.48 \\ 
  \bottomrule
  \end{tabular}
  \end{center}
\end{table*}

\textbf{Practical Benefits} $\quad$ 
% We further highlight the practical benefits from Gumbel-CRF, as is shown in table~\ref{tab:practical}. 
Although our model can be trained on either REINFORCE or Gumbel-CRF, we emphasize that 
training structured variables with REINFORCE is notoriously difficult~\citep{Li2020PosteriorCO}, 
and Gumbel-CRF substantially reduces the complexity. 
Table~\ref{tab:practical} demonstrates this empirically. 
Gumbel-CRF requires fewer hyperparameters to tune, fewer MC samples, less GPU memory, and faster training. 
These advantages would considerably benefit all practitioners with significantly less training time and resource consumption.

\section{Conclusion}
\label{sec:concluson}

In this work, 
we propose a pathwise gradient estimator for sampling from CRFs which exhibits lower variance and more stable training than existing baselines. 
We apply this gradient estimator to the task of text modeling, 
where we use a structured inference network based on CRFs to learn latent templates. 
Just as REINFORCE, models trained with Gumbel-CRF can also learn meaningful latent templates that successfully encode lexical and structural information of sentences, 
thus inducing interpretability and controllability for text generation.
Furthermore, the Gumbel-CRF gives significant practical benefits than REINFORCE, making it more applicable to real-world tasks. 

\section*{Broader Impact}

Generally, this work is about controllable text generation. 
When applying this work to chatbots, one may get a better generation quality. 
This could potentially improve the accessibility~\citep{Corbett2016WhatCI, Reis2018UsingIP} for people who need a voice assistant to use an electronic device, 
e.g. people with visual, intellectual, and other disabilities~\citep{Qidwai2012UbiquitousAV, Balasuriya2018UseOV}. 
However, if not properly tuned, this model may generate improper sentences like fake information, putting the user at a disadvantage. 
Like many other text generation models, if trained with improper data (fake news~\citep{Grinberg2019FakeNO}, words of hatred~\citep{Santos2018FightingOL}), 
a model could generate these sentences as well.
In fact, one of the motivations for controllable generation is to avoid these situations~\citep{wiseman2018learning, Santos2018FightingOL}.
But still, researchers and engineers need to be more careful when facing these challenges.

% \section*{References}
\medskip
% \small
\bibliographystyle{plainnat}
\bibliography{latent_template}  

\newpage
\appendix

\section{CRF Entropy Calculation}
The entropy of the inference network can be calculated by another forward-styled DP algorithm. Algorithm~\ref{algo:ent} gives the datils.

\begin{figure}[t]
  
    \begin{algorithm}[H]
        \centering
        \caption{\footnotesize Linear-chain CRF Entropy}\label{algo:ent}
        \footnotesize
        \begin{algorithmic}[1]
            \State \text{\textbf{Input:} $\Phi(z_{t-1}, z_t, x_t), t \in\{1, .., T\}, \alpha_{1:T}, Z$} 
            \State $\mathcal{H}_1(i) = 0$ \Comment{We assume a deterministic special start state}  
            \For{$t \gets 1, T-1$}
            \State $w_{t + 1}(i, j) = \frac{\Phi(z_t = i, z_{t + 1} = j, x_{t + 1}) \alpha_t(i)}{\alpha_{t + 1}(j)}$
            \State $\mathcal{H}_{t +1}(j) = \sum_i w_{t + 1}(i, j)[\mathcal{H}_t(i) - \log w_{t  +1}(i, j)]$
            \EndFor
            \State $p(z_T = j | x) = \frac{\alpha_T(j)}{\sum_k \alpha_T(k)}$ \Comment{We assume all states ends with a special end state with probability 1.}  
            \State $\mathcal{H} = \sum_j p(z_T = j| x) [\mathcal{H}_T(j) - \log p(z_T = j|x)]$
            \State \textbf{Return}  $\mathcal{H}$ %\Comment{$\hat{z}$ is not differentiable}
        \end{algorithmic}
    \end{algorithm} 
  \end{figure}

\section{PM-MRF}

As noted in the main paper, the baseline estimator PM-MRF also involve in-depth exploitation of the structure of models and gradients, thus being quite competitive. 
Here we give a detailed discussion.

\citet{papandreou2011perturb} proposed the Perturb-and-MAP Random Field, 
 an efficient sampling method for general Markov Random Field. 
Specifically, 
they propose to use the Gumbel noise to perturb each local potential $\Phi_i$ of an MRF, 
then run a MAP algorithm (if applicable) on the perturbed MRF to get a MAP $\hat{z}$.
This MAP $\hat{z}$ from the perturbed $\tilde{\Phi}$ can be viewed as a biased sample from the original MRF.
This method is much faster than the MCMC sampler when an efficient MAP algorithm exists.
Applying to a  CRF, this would mean adding noise to its potential at every step, then run Viterbi:
\begin{subequations}
\begin{align}
  \tilde{\Phi}(z_t = i, x_{t-1}, x_{t}) &= \Phi(z_t, x_{t-1}, x_{t}) + g, g \sim \text{G}(0) \quad \text{for all}\;\; t, i\\
  \hat{z} &= \text{Viterbi}(\tilde{\Phi})
\end{align}
\end{subequations}
However, when tracing back along the Viterbi path, we still get $\hat{z}$ as a sequence of \textit{index}. 
For continuous relaxation, we would like to relax $\hat{z}_t$ to be \textit{relaxed one-hot}, instead of index. 
One natural choice is to use the Softmax function. 
The relaxed back-tracking algorithm is listed in Algorithm \ref{algo:rviterbi}. 
In our experiments, for the PM-MRF estimator, we use $\tilde{z}$ for both forward and back-propagation. 
For the PM-MRF-ST estimator, we use $\hat{z}$ for the forward pass, and $\tilde{z}$ for the back-propagation pass. 

It is easy to verify the PM-MRF is a biased sampler by checking the sample probability of the first step $\hat{z}_1$. 
With the PM-MRF, the biased $z_1$ is essentially from a categorical distribution parameterized by $\pi$ where: 
\begin{equation}
  \log \pi_i = \log \Phi(z_1 = i, x_1)
\end{equation}
With forward-sampling, however, the unbiased $z_1$ should be from the marginal distribution where:
\begin{equation}
  \log \pi_i = \log \beta_1(i) \ne \log \Phi(z_1 = i, x_1) \label{eq:biased}
\end{equation}
Where $\beta$ denote the backward variable from the backward algorithm~\citep{sutton2012introduction}. 
The inequality in equation \ref{eq:biased} shows that PM-MRF gives biased sample.

\begin{figure}[t]
  
  \begin{algorithm}[H]
      \centering
      \caption{\footnotesize Viterbi with Relaxed Back-tracking}\label{algo:rviterbi}
      \footnotesize
      \begin{algorithmic}[1]
          \State \text{\textbf{Input:} $\tilde{\Phi}(z_{t-1}, z_t, x_t), t \in\{1, .., T\}$} 
          \State $s_1(i) = \log \tilde{\Phi}(i, x_1)$
          \For{$t \gets 2, T$}
          \State $s_t(i) = \max_j\{s_{t - 1}(j) + \log \tilde{\Phi}(z_{t-1} = j, z_t = i, x_t)\}$
          \State $b_t(i) = \text{Softmax}_j (s_{t - 1}(j) + \log \tilde{\Phi}(z_{t - 1}=j, z_t = i, x_t))$
          \EndFor
          \State Back-tracking: 
          \State $\tilde{z}_T = \text{Softmax}(s_T)$
          \State $\hat{z}_T = \text{Argmax}(s_T(i))$
          \For{$t \gets T - 1, 1$}
          \State $\hat{z}_{t + 1} = \text{Argmax}_i (\tilde{z}_{t + 1}(i))$
          \State $\tilde{z}_t = b_{t + 1}(\hat{z}_{t +1})$
          \EndFor
          \State \textbf{Return}  $\hat{z}, \tilde{z}$ %\Comment{$\hat{z}$ is not differentiable}
      \end{algorithmic}
  \end{algorithm} 
\end{figure}

\section{Theoretical Comparison of Gradient Structures}
We compare the detailed structure of gradients of each estimator. 
% We assume a linear-chain CRF inference model $q_\phi(z|x)$.
We denote $f(x_{1:n}, z_{1:n}) = \log p_\theta(x_{1:n}, z_{1:n})$. 
We use $\hat{z}$ to denote unbiased hard sample, $\tilde{z}$ to denote soft sample coupled with $\hat{z}$, 
$\hat{z}'$ to denote biased hard sample from the PM-MRF, $\tilde{z}'$ to denote soft sample coupled with $\hat{z}'$ output by the relaxed Viterbi algorithm. 
We use $w_{1:n}$ to denote the ``emission'' weights of the CRF. 
The gradients of all estimators are:

\begin{align}
   \nabla_\phi \mathcal{L}_{\text{REINFORCE}}  &\approx \sum_t \underbrace{f(x_{1:n}, \hat{z}_{1:n})}_{\text{reward term}} \cdot \underbrace{\nabla_\phi \log q_\phi(\hat{z}_t | \hat{z}_{t - 1}, x)}_{\text{stepwise term}} \label{eq:reinforce}\\ 
  \nabla_\phi \mathcal{L}_{\text{Gumbel-CRF-ST}} &\approx \sum_t \underbrace{\nabla_{\tilde{z}_t} f(x_{1:n}, \hat{z}_{1:n})}_{\text{pathwise term}} \odot \underbrace{\nabla_\phi \tilde{z}_t (\hat{z}_{t + 1}, w_{1:n}, \epsilon_t)}_{\text{stepwise term}} \label{eq:gumbelcrf} \\ 
  \nabla_\phi \mathcal{L}_{\text{PM-MRF-ST}} &\approx \sum_t \underbrace{\nabla_{\tilde{z}'_t} f(x_{1:n}, \hat{z}'_{1:n})}_{\text{pathwise term}} \odot \underbrace{\nabla_\phi \tilde{z}'_t (\hat{z}'_{t + 1},  w_{1:n}, \epsilon_t)}_{\text{stepwise term}} \label{eq:pmmrf}
\end{align}
In equation~\ref{eq:reinforce}, we decompose $q(z | x)$ with its markovian property, leading to a summation over the chain where the same reward $f$ is distributed to all steps. 
Equations~\ref{eq:gumbelcrf} and~\ref{eq:pmmrf} use the chain rule to get the gradients. 
$\nabla_{\tilde{z}_t} f(x_{1:n}, \hat{z}_{1:n})$ denotes the gradient of $f$ evaluated on hard sample $\hat{z}_{1:n}$ and taken w.r.t. soft sample $\tilde{z}_t$. 
$\nabla_\phi \tilde{z}_t (\hat{z}_{t + 1}, w_{1:n}, \epsilon_t)$ denotes the Jacobian matrix of $\tilde{z}_t$ (note $\tilde{z}_t$ is a vector) taken w.r.t. the parameter $\phi$ (note $\phi$ is also a vector, so taking gradients of $\tilde{z}_t$ w.r.t. $\phi$ gives a Jacobian matrix). 
Consequently $\odot$ is a special vector-matrix summation which result in a vector (note this is different with equation~\ref{eq:reinforce} since the later is a scalar-vector product). 
We further use $\tilde{z}_t (\hat{z}_{t + 1}, w_{1:n}, \epsilon_t)$ to denote that $\tilde{z}_t$ is a function of the previous hard sample $\hat{z}_{t + 1}$, all CRF weights $w_{1:n}$, and the local Gumbel noise $\epsilon_t$. 
Similar notation applies to equation~\ref{eq:pmmrf}. 

All gradients are formed as a summation over the steps. 
Inside the summation is a scalar-vector product or a vector-matrix product. 
The REINFORCE estimator can be decomposed with a reward term and a ``stepwise'' term, where the stepwise term comes from the ``transition'' probability.
The Gumbel-CRF and PM-MRF estimator can be decomposed with a pathwise term, where we take gradient of $f$ w.r.t. each sample step $\tilde{z}_t$ or $\tilde{z}'_t$, and a ``stepwise'' term where we take Jacobian w.r.t. $\phi$. 

To compare the three estimators, we see that:
\begin{itemize}
  \item \textbf{Using hard sample $\hat{z}$}. like REINFORCE, Gumbel-CRF-ST use hard sample $\hat{z}$ for the forward pass, as indicated by the term $f(x_{1:n}, z_{1:n})$
  \begin{itemize}
    \item The advantage of using the hard sample is that one can use it to best explore the search space of the inference network, i.e. to search effective latent codes using Monte Carlo samples.  
    \item Gumbel-CRF-ST perserves the same advantage as REINFORCE, while PM-MRF-ST cannot fully search the space because its sample $\hat{z}'$ is biased. 
  \end{itemize}
  \item \textbf{Coupled sample path}. The soft sample $\tilde{z}_t$ of Gumbel-CRF-ST is based on the hard, exact sample path $\hat{z}_{t + 1}$, as indicated by the term $\tilde{z}_t (\hat{z}_{t + 1}, w_{1:n}, \epsilon_t)$. 
  \begin{itemize}
    \item The coupling of hard $\hat{z}$ and soft $\tilde{z}$ is ensured by our Gumbelized FFBS algorithm by applying gumbel noise $\epsilon_t$ to each transitional distribution $\tilde{z}_t = \text{Softmax}(\log q(z_t | \hat{z}_{t + 1}, x) + \epsilon_t$). 
    \item Consequently, we can recover the hard sample with the Argmax function $\hat{z}_t = \text{Argmax}(\tilde{z}_t)$. 
    \item This property allows us the use continuous relaxation to allow pathwise gradients $\nabla_\phi \tilde{z}_t (\hat{z}_{t + 1}, w_{1:n}, \epsilon_t)$ without losing the advantage of using hard exact sample $\hat{z}$. 
    \item PM-MRF with relaxed Viterbi also has this advantage of continuous relaxation, as shown by the term $\nabla_\phi \tilde{z}'_t (\hat{z}'_{t + 1}, w_{1:n}, \epsilon_t)$, but it does not have the advantage of using unbiased sample since $\hat{z}'_{t + 1}$ is biased.
  \end{itemize}
  \item \textbf{``Fine-grained'' gradients}. The stepwise term $\nabla_\phi \log q_\phi(\hat{z}_t | \hat{z}_{t - 1}, x)$ in the REINFORCE estimator is scaled by the same reward term $f(x_{1:n}, \hat{z}_{1:n})$, while the stepwise term $\nabla_\phi \tilde{z}_t (\hat{z}_{t + 1}, w_{1:n}, \epsilon_t)$ in the rest two estimators are summed with different pathwise terms $\nabla_{\tilde{z}_t} f(x_{1:n}, \hat{z}_{1:n})$. 
  \begin{itemize}
    \item To make REINFORCE achieve similar ``fine-grained'' gradients for each steps, the reward function (generative model) $f$ must exhibit certain structures that make it decomposible. This is not always possible, and one always need to manually derive such decomposition. 
    \item The fine-grained gradients of Gumbel-CRF is agnostic with the structure of the generative model. No matter what $f$ is, the gradients decompose automatically with AutoDiff libraries. 
  \end{itemize}
\end{itemize}

\section{Experiment Details}
\subsection{Data Processing}
For the \texttt{E2E} dataset, we follow similar processing pipeline as \citet{wiseman2018learning}. 
Specifically, given the key-value pairs and the sentences, we substitute each value token in the sentence with its corresponding key token. 
For the \texttt{MSCOCO} dataset, we follow similar processing pipeline as \citet{Fu2019ParaphraseGW}. 
Since the official test set is not publically available, we use the same training/ validation/ test split as \citet{Fu2019ParaphraseGW}. 
We are unable to find the implementation of \citet{Liu2019UnsupervisedPB}, thus not sure their exact data processing pipeline, 
making our results of unsupervised paraphrase generation not strictly comparable with theirs. 
However, we have tested different split of the validation dataset, 
and the validation performance \textit{does not change significantly with the split}. 
This indicates that although not strictly comparable, we can assume their testing set is just another random split, 
and their performance should not change much under our split. 

\subsection{Model Architecture}

For the inference model, we use a bi-directional LSTM to predict the CRF emission potentials. 
% The embedding size and the state size are both 300. 
The dropout ratio is 0.2. 
The number of latent state of the CRF is 50.
%  and the embedding size for the latent state is also 300. 
The decoder is a uni-directional LSTM model. 
We perform attention to the BOW, and also let the decoder to copy~\citep{gu2016incorporating} from the BOW.  
%  and its state size is 300
For text modeling and data-to-text, we set the number of LSTM layers to 1 (both encoder and decoder), and the hidden state size to 300.
This setting is comparable to \citep{wiseman2018learning}.
For paraphrase generation, we set the number of LSTM layers (both encoder and decoder) to 2, and the hidden state size to 500.
This setting is comparable to \citep{Fu2019ParaphraseGW}. 
The embedding size for the words and the latent state is the same as the hidden state size in both two settings.

\subsection{Hyperparameters, Training and Evaluation Details}

\paragraph{Hyperparameters} 
For the score function estimators, we conduct more than 40 different runs searching for the best hyperparameter and architecture, 
and choose the best model according to the validation performance. 
The hyperparameters we searched include: 
(a). number of MC sample (3, 5) 
(b). value of the constant baseline (0, 0.1, 1.0)
(c). $\beta$ value ($5\times10^{-6}, 10^{-4}, 10^{-3}$)
(d). scaling factor of the surrogate loss of the score function estimator (1, $10^2$, $10^4$).
For the reparameterized estimators, we conduct more than 20 different runs for architecture and hyperparameter search. 
The hyperparameters we searched include:
(a). the template in Softmax (1.0, 0.01)
(b). $\beta$ value ($5\times10^{-6}, 10^{-4}, 10^{-3}$).
Other parameter/ architecture we consider include:
(a). number of latent states (10, 20, 25, 50)
(b). use/ not use the copy mechanism
(c). dropout ratio
(d). different word drop schedule. 
Although we considered a large range of hyperparameters, we have not tested all combinations.
For the settings we have tested, all settings are repeated 2 times to check the sensitivity under different random initialization. 
If we find a hyperparameter setting is sensitive to initialization, we run this setting 2 more times and choose the best. 

\paragraph{Training} 
We find out the convergence of score-function estimators are generally less stable than the reparameterized estimators, 
they are: (a). more sensitive to random initialization (b). more prone to converging to a collapsed posterior. 
For the reparameterized estimators, the ST versions generally converge faster than the original versions. 
% However, ST versions are also prone to converging to a worse local optimal than the original versions. 

% \paragraph{Evaluation}
% For estimating the ELBO, we use the exact samples from the CRF model with the FFBS algorithm. 
% The number of samples is 5. 
% We have tested a larger number of samples up to 50, and the results do not change significantly. 
% We have also estimated the reconstruction NLL and the marginal likelihood with importance sampling, and the ranking of the models also does not change much.
% Importantly, for all the metrics we tested, the Gumbel-CRF (original or ST version) outperforms all other baselines, showing the effectiveness of our approach. 

% \end{CJK*}
\end{document}